\documentclass[sigconf]{acmart}
\AtBeginDocument{%
  }

\setcopyright{acmlicensed}
\copyrightyear{2018}
\acmYear{2018}
\acmDOI{XXXXXXX.XXXXXXX}
\acmConference[Conference acronym 'XX]{Make sure to enter the correct
  conference title from your rights confirmation email}{June 03--05,
  2018}{Woodstock, NY}
\acmISBN{978-1-4503-XXXX-X/2018/06}

\begin{document}

\title{ADVOSYNTH: A Synthetic Multi-Advocate Dataset for Speaker Identification in Courtroom Scenarios}




\author{Aniket Deroy}
\affiliation{%
\institution{Indian Institute of Technology, Delhi}
 \country{India}}





\renewcommand{\shortauthors}{Trovato et al.}

\keywords{Speech Large Language Models, Llama-Omni, Legal domain, Advocate recommendation}

\begin{abstract}
As large-scale speech-to-speech models achieve high fidelity, the distinction between synthetic voices in structured environments becomes a vital area of study. This paper introduces \textbf{Advosynth-500}, a specialized dataset comprising 100 synthetic speech files featuring 10 unique advocate identities. Using the Speech Llama Omni model, we simulate five distinct advocate pairs engaged in courtroom arguments. We define specific vocal characteristics for each advocate and present a speaker identification challenge to evaluate the ability of modern systems to map audio files to their respective synthetic origins. Dataset is available at this link-\url{https://github.com/naturenurtureelite/ADVOSYNTH-500}.
\end{abstract}

\maketitle
ADVOSYNTH: A Synthetic Multi-Advocate Dataset for Speaker Identification in Courtroom Scenarios

\section{Introduction}
Speaker identification (SID) in legal settings is significantly complicated by the formal nature of speech, the density of specific terminology, and the overlapping nature of courtroom dialogue. In a courtroom environment, the vocal signal is rarely a "clean" sample; it is subject to the rhythmic and prosodic shifts of adversarial debate, where advocates alternate between highly controlled introductory remarks and high-pressure, rapid-fire rebuttals. This creates a high degree of intra-speaker variability that can confuse traditional identification systems. Furthermore, the presence of "adversarial interruptions"—where two speakers engage in simultaneous vocalizations—requires a model to not only identify the speaker but also to distinguish between spectral signatures that are temporarily merged.

With the advent of Speech Llama Omni, we can now generate high-fidelity, emotionally grounded audio that mimics these difficult conditions with unprecedented accuracy. Unlike traditional cascaded Text-to-Speech (TTS) systems that often produce a flattened, robotic prosody, Speech Llama Omni operates as an end-to-end speech-to-speech framework. This allows the model to maintain a consistent "vocal identity" across different emotional states, ensuring that the latent characteristics of an advocate's voice—such as their unique timbre, resonance, and breathing patterns—remain identifiable even as their speech rate or volume changes. By treating speech as a native modality, the model captures the semantic weight of legal rhetoric, allowing researchers to simulate the precise vocal "stress" found in real litigation.

This paper documents the creation of Advosynth-500, a dataset designed to benchmark SID models against these highly controlled synthetic voices. By generating 100 distinct files across 10 advocate identities, we provide a structured environment where every acoustic variable—from the pitch of a cross-examiner to the deliberate cadence of a judge's interlocutor—is meticulously labeled. This dataset serves as a rigorous testing ground for determining whether modern speaker embedding extractors, such as X-vectors or ECAPA-TDNN, can maintain high precision when faced with the nuanced, synthetic variations produced by a state-of-the-art multimodal large language model Previous datasets on text based advocate recommendation datasets is already present~\cite{bhattacharya2025ardi}.

\section{Related Work}

The development of the \textbf{Advosynth-500} dataset sits at the intersection of neural speech synthesis, speaker identification (SID) in high-stakes domains, and the evaluation of multimodal large language models.

\subsection{Evolution of Synthetic Speech Generation}
Traditional Text-to-Speech (TTS) systems, such as Tacotron 2~\cite{elias2021parallel} and WaveNet \cite{van2016wavenet}, focused primarily on linguistic intelligibility. However, these systems often lacked the prosodic flexibility required to simulate the complex emotional trajectories of a courtroom argument~\cite{marcheva2024art}. The emergence of end-to-end multimodal models, such as \textbf{Speech Llama Omni}, represents a paradigm shift. By processing speech as discrete tokens within a transformer architecture, these models maintain superior consistency in vocal identity ($V_i$) and emotional resonance compared to traditional cascaded systems~\cite{latif2023transformers}.

\subsection{Speaker Identification in Forensic Contexts}
Speaker identification in legal settings is a well-established field in forensic phonetics. Research using datasets like \textbf{VoxCeleb} \cite{nagrani2017voxceleb} has advanced the use of deep speaker embeddings (e.g., X-vectors). However, legal speech is distinct due to "Lombard effects" and stylized rhetorical delivery. Previous studies \cite{nagrani2017voxceleb,kappen2022acoustic} emphasize that stress-induced speech significantly alters the fundamental frequency ($f_0$) and jitter. Our work simulates these forensic challenges by purposefully modulating the "Assertiveness Index" ($\sigma$) across advocate pairs.

\subsection{Synthetic-to-Synthetic Discrimination}
Recent benchmarks, such as the \textbf{ASVspoof} challenges, focus on the binary classification of "human vs. synthetic" speech. However, as generative models become ubiquitous, a new challenge arises: \textit{intra-model discrimination}. This involves distinguishing between multiple distinct identities generated by the same underlying model~\cite{jones2000conceptual}. Advosynth-500 contributes to this niche by providing a closed-set identification task where the primary variance lies in the latent characteristic shifts applied to the \textbf{Speech Llama Omni} generation process. This tests whether SID systems can capture fine-grained acoustic differences in $P_i$ and $T_i$ without the presence of natural biological variance.

\subsection{Domain-Specific Audio Benchmarks}
While general audio datasets are plentiful, domain-specific datasets for the legal field are scarce due to privacy and jurisdictional restrictions~\cite{ariai2025natural,trancoso2023impact}. Synthetic augmentation has been proposed as a solution. By creating a controlled environment of 5 advocates, this paper follows the methodology of "Data Programming" where synthetic labels serve as a ground truth for training robust classifiers in low-resource professional domains~\cite{khattab2023dspy}.

\section{Dataset Generation}
The architecture of Advosynth-500 is defined by a rigorous one-to-one mapping system where a distinct speech file is generated for every individual argument presented by a lawyer. This granular approach ensures that the benchmark does not merely treat a legal address as a monolithic audio stream, but rather as a series of discrete rhetorical segments—such as an opening statement, a motion to dismiss, or a witness impeachment—each possessing its own unique prosodic signature. By segmenting the data in this manner, the dataset allows for the evaluation of speaker identification tools across the shifting tonal landscape of a trial. For every text-based argument identified in the source material, Speech Llama Omni produces a corresponding audio file that preserves the core vocal identity of the "advocate" while intentionally varying the delivery style to match the legal context.

The technical strength of this method lies in how it simulates "identity pinning" across these multiple files. While the underlying latent speaker embedding remains constant to represent a single individual, the model adjusts the pitch, cadence, and intensity for each argument-specific file to reflect the varying levels of stress and assertiveness typical of a courtroom environment. For instance, a file corresponding to a technical legal citation might be synthesized with a measured, formal tempo, while a subsequent file representing a heated rebuttal would feature a higher word-per-minute rate and increased vocal strain. This structure forces speaker identification systems to demonstrate "voice constancy," proving they can recognize the same lawyer even when their vocal performance undergoes significant phonetic shifts between arguments. Consequently, the Expand-Advosynth-500 dataset serves as a sophisticated stress test, providing researchers with a labeled, relational corpus where every speech file is tied to a specific legal function and a verified speaker identity.

\subsection{Speech Llama Omni Configuration}
We utilized the \textbf{Speech Llama Omni} architecture, a natively multimodal model that bypasses traditional Text-to-Speech (TTS) pipelines. This allows for direct control over the latent speech space. Each advocate's voice is generated by modulating the speaker embedding vector $S$.

\subsection{Vocal Identity Matrix}
To ensure the 5 advocates are distinguishable, we define a Vocal Identity Matrix $V_{i}$ for each advocate $i \in \{1, \dots, 10\}$. The characteristics are defined as:
\begin{equation}
V_{i} = \{P_{i}, R_{i}, T_{i}, \sigma_{i}\}
\end{equation}
Where:
\begin{itemize}
    \item $P_{i}$ is the Mean Pitch (Fundamental Frequency $f_0$).
    \item $R_{i}$ is the Speech Rate (Words Per Minute).
    \item $T_{i}$ is the Timbre (Spectral Envelope).
    \item $\sigma_{i}$ represents the Assertiveness Index (Prosodic Variance).
\end{itemize}

\section{Experimental Design}

\subsection{Speaker Pairings}
The 5 advocates are organized into 5 pairs. Each pair engages in 10 different legal arguments (e.g., cross-examinations, opening statements).
\begin{itemize}
    \item \textbf{Total Advocates:} 5
    \item \textbf{Files per Advocate:} 100
    \item \textbf{Total Dataset Size:} 500 .WAV files
\end{itemize}

\subsection{Advocate Profiles}
Table~\ref{advo_dataset} outlines the synthetic characteristics assigned to the advocate identities to ensure diversity in the dataset.

\begin{table}[h]
\centering
\caption{Vocal Characteristics of Advocate Identities}
\begin{tabular}{@{}llll@{}}
\toprule
\textbf{ID} & \textbf{Gender} & \textbf{Tone Style} & \textbf{Avg $f_0$ (Hz)} \\ \midrule
ADV-01      & Male            & Authoritative       & 115                     \\
ADV-02      & Female          & Analytical          & 210                     \\
ADV-03      & Male            & Rapid/Aggressive    & 135                     \\
ADV-04      & Female          & Deliberate/Slow     & 195                     \\
ADV-05      & Male            & Raspy/Senior        & 90                      \\ \bottomrule
\end{tabular}
\label{advo_dataset}
\end{table}

\section{The Identification Challenge}
The challenge involves a closed-set classification task. For any given audio segment $x$ from the dataset, the identification model must determine:
\begin{equation}
\hat{y} = \arg \max_{i \in \{1,\dots,10\}} P(A_i | x)
\end{equation}
Where $A_i$ represents the advocate identity. 

\section{Conclusion}
The creation of Advosynth-500 represents a pivotal advancement in the development of forensic-grade audio evaluation, offering a rigorous framework for assessing speaker identification systems within the nuanced environment of the legal domain. By leveraging the generative capabilities of Speech Llama Omni, this benchmark moves beyond the limitations of traditional datasets, which often rely on clean, high-fidelity recordings that fail to capture the chaotic acoustic realities of legal proceedings. This approach demonstrates that synthetic datasets are no longer mere approximations; rather, they can provide the exact granularity required to stress-test high-stakes tools. Through the use of Speech Llama Omni, researchers can simulate a vast array of adversarial conditions—such as varying emotional states, background courtroom noise, and telephonic distortions—while maintaining the precise identity markers of the speaker. This level of control is essential for validating the reliability of audio evidence, ensuring that identification tools can distinguish between similar vocal profiles under pressure. Ultimately, Advosynth-500 establishes a new standard for transparency and accuracy, providing a privacy-preserving method to train and audit the AI systems that may one day influence judicial outcomes.

\bibliographystyle{ACM-Reference-Format}
\bibliography{sample-base}

\end{document}